\documentclass[10pt,twocolumn,letterpaper]{article}

\usepackage{iccv}
\usepackage{times}
\usepackage{epsfig}
\usepackage{graphicx}
\usepackage{amsmath}
\usepackage{amssymb}
\usepackage{multirow}
\usepackage{subfigure}

\usepackage[pagebackref=true,breaklinks=true,letterpaper=true,colorlinks,bookmarks=false]{hyperref}

\iccvfinalcopy 

\def\iccvPaperID{1996} 

\ificcvfinal\fi
\begin{document}

\title{Interleaved Group Convolutions for Deep Neural Networks}

\author{
Ting Zhang$^1$
\quad
Guo-Jun Qi$^2$
\quad
Bin Xiao$^1$
\quad
Jingdong Wang$^1$
\\
$^1$Microsoft Research
\quad
$^2$University of Central Florida\\
{\tt\small \{tinzhan, Bin.Xiao, jingdw\}@microsoft.com}
\quad
{\tt\small guojun.qi@ucf.edu }
}

\maketitle
\thispagestyle{empty}

\begin{abstract}
In this paper,
we present a simple and modularized neural network architecture,
named interleaved group convolutional neural networks (IGCNets).
The main point lies in a novel building block,
a pair of two successive interleaved group convolutions:
primary group convolution and secondary group convolution.
The two group convolutions are complementary:
(i) the convolution on each partition in primary group convolution
is a spatial convolution, while on each partition
in secondary group convolution,
the convolution is a point-wise convolution;
(ii) the channels in the same secondary partition
come from different primary partitions.
We discuss one representative advantage:
Wider than a regular convolution
with the number of parameters
and the computation complexity preserved.
We also show that
regular convolutions, group convolution with summation fusion,
and the Xception block
are special cases of interleaved group convolutions.
Empirical results
over standard benchmarks, CIFAR-$10$, CIFAR-$100$, SVHN and ImageNet
demonstrate that our networks are more efficient
in using parameters and computation complexity
with similar or higher accuracy.
\end{abstract}

\section{Introduction}
Architecture design in deep convolutional neural networks
has been attracting increasing interests.
The basic design purpose is efficient
in terms of computation and parameter
with high accuracy.
Various design dimensions have been considered,
ranging from small kernels~\cite{IoannouRSCC15, SzegedyVISW16, SzegedyIVA17, Chollet16a, IoannouRCC16},
identity mappings~\cite{HeZRS16}
or general multi-branch structures~\cite{WangWZZ16, ZhaoWLTZ16, LarssonMS16a, SzegedyLJSRAEVR15, SzegedyVISW16, SzegedyIVA17}
for easing the training of very deep networks,
and multi-branch structures for increasing the width~\cite{SzegedyLJSRAEVR15, Chollet16a, IoannouRCC16}.

Our interest is to reduce the redundancy of convolutional kernels.
The redundancy comes from two extents:
the spatial extent and
the channel extent.
In the spatial extent,
small kernels are developed,
such as $3 \times 3$,
$3 \times 1$, $1\times 3$~\cite{SzegedyVISW16, SimonyanZ14a, JaderbergVZ14, MamaletG12, JinDC14}.
In the channel extent,
group convolutions~\cite{ZhaoWLTZ16, XieGDTH16} and
channel-wise convolutions or separable filters~\cite{SifreM14, Chollet16a, IoannouRCC16},
have been studied.
Our work belongs to the kernel design
in the channel extent.

\begin{figure*}[t]
\begin{center}
\includegraphics[width = 1\textwidth, clip]{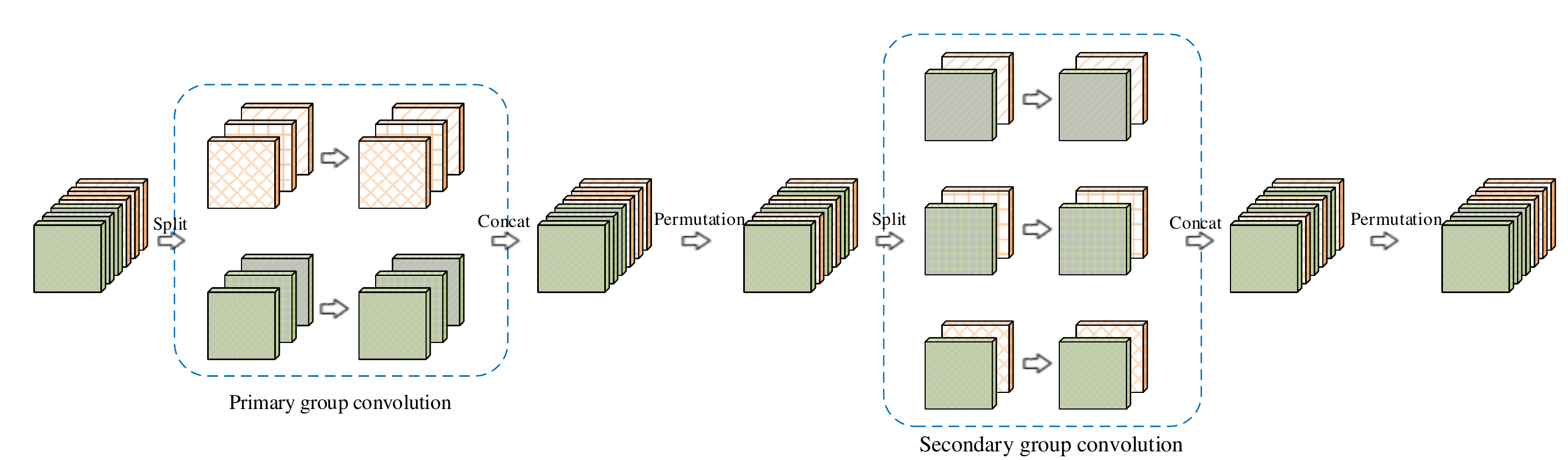}
\end{center}
   \caption{Illustrating the interleaved group convolution,
   with $L=2$ primary partitions
   and $M=3$ secondary partitions.
   The convolution for each primary partition in primary group convolution
   is spatial.
   The convolution for each secondary partition in secondary group convolution
   is point-wise ($1 \times 1$).
   Details are given in Section~\ref{sec:pdgc}.}
\label{fig:pdgc}
\vspace{-.3cm}
\end{figure*}

In this paper,
we present a novel network architecture,
which is a stack of interleaved group convolution (IGC) blocks.
Each block contains two group convolutions:
primary group convolution and secondary group convolution,
which are conducted on primary and secondary partitions, respectively.
The primary partitions are obtained
by simply splitting input channels,
e.g., $L$ partitions with each containing $M$ channels,
and there are $M$ secondary partitions,
each containing $L$ channels
that lie in different primary partitions.
The primary group convolution
performs the spatial convolution over each primary partition
\emph{separately},
and the secondary group convolution performs a $1\times 1$ convolution
(point-wise convolution)
over each secondary partition,
\emph{blending} the channels across partitions outputted by primary group convolution.
Figure~\ref{fig:pdgc} illustrates the interleaved group convolution block.

It is known that a group convolution
is equivalent to a regular convolution
with sparse kernels:
there is no connections across the channels
in different partitions.
Accordingly, an IGC block
is equivalent to a regular convolution
with the kernel composed from
the product of two sparse kernels,
resulting in a dense kernel.
We show that
under the same number of parameters/computation complexity,
an IGC block
(except the extreme case that the number of primary partitions, $L$, is $1$)
is wider than a regular convolution with the spatial kernel
size same to that of primary group convolution.
Empirically,
we also observe that
a network built by stacking IGC blocks
under the same computation complexity and the same number of parameters
performs better
than the network with regular convolutions.

We study the relations with existing related modules.
(i) The regular convolution
and group convolution with summation fusion~\cite{XieGDTH16, ZhaoWLTZ16, WangWZZ16},
are both interleaved group convolutions,
where the kernels are in special forms and are fixed in secondary group convolution.
(ii) An IGC block in the extreme case
where there is only one partition in
the secondary group convolution, is very close to Xception~\cite{Chollet16a}.

Our main contributions are summarized as follows.
\begin{itemize}
\setlength\itemsep{0em}
  \item We present a novel building block, interleaved group convolutions,
  which is efficient in parameter and computation.
  \item We show that the proposed building block
  is wider than a regular group convolution
  while keeping the network size and computational complexity,
  showing superior empirical performance.
  \item We discuss the connections to
  regular convolutions, the Xception block~\cite{Chollet16a}, and group convolution with summation fusion,
  and show that they are specific instances of interleaved group convolutions.
\end{itemize}

\section{Related Works}
\noindent\textbf{Group convolutions and multi-branch.}
Group convolution is used in AlexNet~\cite{KrizhevskySH12}
for distributing the model over two GPUs
to handle the memory issue.
The channel-wise convolutions used in the separable
convolutions~\cite{SifreM14},
is an extreme case
of group convolutions,
in which each partition contains only one channel.

The multi-branch architecture
can be viewed as an extension of group convolutions
by generalizing the convolution transformation on each partition,
e.g., different number of convolution layers on different partitions,
such as Inception~\cite{SzegedyLJSRAEVR15}, deeply-fused nets~\cite{WangWZZ16},
a simple identity connection~\cite{HeZRS16}, and so on.
Summation~\cite{XieGDTH16, WangWZZ16}, average~\cite{LarssonMS16a},
and convolution operations~\cite{SzegedyLJSRAEVR15, Chollet16a} following concatenation
are often adopted to blend the outputs.
Our approach further improves parameter efficiency
and adopts primary and secondary group convolutions,
where secondary group convolution acts as a role
of blending the channels
outputted by primary group convolution.

\vspace{.1cm}
\noindent\textbf{Sparse convolutional kernels.}
Sparse convolution kernels have already been embedded
into convolutional neural networks:
the convolution filters usually
have limited~\emph{spatial} extent.
Low-rank filters~\cite{IoannouRSCC15, JaderbergVZ14, MamaletG12}
learn small basis filters,
further sparsifying the connections.
Channel-wise random sparse connection~\cite{ChangpinyoSZ17}
sparsifies the filters
in the~\emph{channel} extent that every output channel is connected
to a small subset of input channels.
There are some works introducing regularizations,
such as structured sparsity regularizer~\cite{LiKDSG16, WenWWCL16},
$\ell_1$ or $\ell_2$ regularization~\cite{HanMD15, HanPTD15}
on the kernel weights.

Our approach also sparsifies kernels
in the~\emph{channel} extent,
and differently,
we use structured sparse connections in primary group convolution:
both input and output convolutional channels
are split to disjoint partitions
and each output partition
is connected
to a single input partition
and vice versa.
In addition,
we use secondary group convolution,
another structured sparse filters,
so that there is a path connecting each channel outputted by secondary group convolution
to each channel fed into primary group convolution.
Xception~\cite{Chollet16a},
which is shown to be
more efficient than Inception~\cite{IoffeS15},
is close to our approach,
and we show that it is a special case of our IGC block.

\vspace{.1cm}
\noindent\textbf{Decomposition.}
Tensor decomposition over each layer's kernel (tensor)
is widely-used to reduce redundancy
of neural networks and compress/accelerate them.
Tensor decomposition usually
finds a low-rank tensor to approximate the tensor
through decomposition along
the spatial dimension~\cite{DentonZBLF14, JaderbergVZ14},
or the input and output channel dimensions~\cite{DentonZBLF14, KimPYCYS15, JaderbergVZ14}.
Rather than
compressing previously-trained networks
by approximating a convolution kernel
using the product of two sparse kernels corresponding to our primary and secondary group convolutions,
we train our network from scratch
and show that our network can improve parameter efficiency and classification accuracy.

\section{Our Network}
\subsection{Interleaved Group Convolutions}
\label{sec:pdgc}
\noindent\textbf{Definition.}
Our building block is based on group convolution,
which is a method
of dividing the input channels into several partitions
and performing a regular convolution
over each partition separately.
A group convolution
can be viewed as a regular convolution
with a sparse block-diagonal convolution kernel,
where each block corresponds to a partition of channels
and there are no connections across the partitions.

Interleaved group convolutions
consist of two group convolutions,
primary group convolution
and secondary group convolution.
An example is shown in Figure~\ref{fig:pdgc}.
We use primary group convolutions
to handle spatial correlation,
and adopt spatial convolution kernels,
e.g., $3 \times 3$, widely-used in state-of-the-art networks~\cite{HeZRS16, SimonyanZ14a}.
The convolutions are performed
over each partition of channels~\emph{separately}.
We use secondary group convolution
to \emph{blend} the channels across partitions
outputted by primary group convolution
and simply adopt $1 \times 1$ convolution kernels.

\vspace{.1cm}
\noindent\textbf{Primary group convolutions.}
Let $L$ be the number of partitions,
called primary partitions,
in primary group convolution.
We choose that each partition
contains the same number ($M$) of channels.
We simplify the discussion
and present the group convolution
over a single spatial position,
and the formulation is easily obtained
for all spatial positions.
The primary group convolution
is given as follows,
\begin{align}
\begin{bmatrix}
       \mathbf{y}_{1} \\[0.3em]
       \mathbf{y}_{2} \\[0.3em]
       \vdots \\[0.3em]
       \mathbf{y}_{L}
     \end{bmatrix} =
     \begin{bmatrix}
     \mathbf{W}_{11}^p & \boldsymbol{0} &  \boldsymbol{0} &  \boldsymbol{0} \\[0.3em]
     \boldsymbol{0} & \mathbf{W}_{22}^p  & \boldsymbol{0} & \boldsymbol{0} \\[0.3em]
     \vdots & \vdots & \ddots  & \vdots \\[0.3em]
     \boldsymbol{0} & \boldsymbol{0} & \boldsymbol{0} & \mathbf{W}_{LL}^p
     \end{bmatrix}
     \begin{bmatrix}
       \mathbf{z}_{1} \\[0.3em]
       \mathbf{z}_{2} \\[0.3em]
       \vdots \\[0.3em]
       \mathbf{z}_{L}
     \end{bmatrix}.
     \label{eqn:kgroupconvolution}
\end{align}
Here $\mathbf{z}_l$ is a $(MS)$-dimensional vector,
with $S$ being the kernel size,
e.g., $9$ for $3\times 3$ kernels,
and it is formed from
the $S$ (e.g., $3 \times 3$) responses around this spatial position
for all the channels in this partition.
$\mathbf{W}_{ll}^p$ corresponds to the convolutional kernel
in the $l$th partition,
and is a matrix of size $M \times (MS)$.
Let $\mathbf{x} = [\mathbf{z}_1^\top~\mathbf{z}_2^\top~\dots~\mathbf{z}_L^\top]^\top$\
represent the input of primary group convolution.

\vspace{.1cm}
\noindent\textbf{Secondary group convolutions.}
Our approach permutes
the channels outputted by primary group convolution,
$\{\mathbf{y}_1, \mathbf{y}_2, \dots, \mathbf{y}_L\}$,
into $M$ secondary partitions
with each partition consisting of $L$ channels,
such that the channels in the same secondary partition
come from different primary partitions.
We adopt a simple scheme
to form the secondary partitions:
the $m$th secondary partition is composed
of the $m$th output channel from each primary partition,
\begin{align}
\bar{\mathbf{y}}_m
= [y_{1m}~y_{2m}~\dots~y_{Lm}]^\top
= \mathbf{P}_m^\top \mathbf{y},~\bar{\mathbf{y}} = \mathbf{P}^\top \mathbf{y}.
\end{align}
Here, $\bar{\mathbf{y}}_m$
corresponds to the $m$th secondary partition,
$y_{lm}$ is the $m$th element of $\mathbf{y}_{l}$,
$\bar{\mathbf{y}} = [\bar{\mathbf{y}}_1^\top~\bar{\mathbf{y}}_2^\top~\dots~\bar{\mathbf{y}}_M^\top]^\top$.
$\mathbf{y} = [\mathbf{y}_1^\top~\mathbf{y}_2^\top~\dots~\mathbf{y}_L^\top]^\top$.
$\mathbf{P}$ is the permutation matrix,
and
$\mathbf{P} = [\mathbf{P}_1~\mathbf{P}_2~\dots~\mathbf{P}_M]$.

The secondary group convolution is performed
over the $M$ secondary partitions:
\begin{align}
\bar{\mathbf{z}}_m = \mathbf{W}_{mm}^d \bar{\mathbf{y}}_m,
\label{eqn:dualgroupconvolution}
\end{align}
where $\mathbf{W}_{mm}^d$
corresponds to the $1\times 1$ convolution kernel
of the $m$th secondary partition,
and is a matrix of size $L \times L$.
The channels outputted by secondary group convolution
are permuted back
to the primary form
as the input of the next
interleaved group convolution block.
The $L$ permuted-back partitions
are given as follows,
$\{\mathbf{x}_1', \mathbf{x}_2', \dots, \mathbf{x}_L'\}$,
and
\begin{align}
\mathbf{x}_l'
= [\bar{z}_{1l}~\bar{z}_{2l}~\dots~\bar{z}_{Ml}]^\top,~
\mathbf{x}' = \mathbf{P} \bar{\mathbf{z}},
\end{align}
where $\bar{\mathbf{z}} = [\bar{\mathbf{z}}_1^\top~\bar{\mathbf{z}}_2^\top~\dots~\bar{\mathbf{z}}_M^\top]^\top$.

In summary,
an interleaved group convolution block is formulated as
\begin{align}
\mathbf{x}'=\mathbf{P}\mathbf{W}^d\mathbf{P}^\top\mathbf{W}^p\mathbf{x},
\label{eqn:PDGCformulation}
\end{align}
where $\mathbf{W}^p$
and $\mathbf{W}^d$
are block-diagonal matrices:
$\mathbf{W}^p = \operatorname{diag}(\mathbf{W}_{11}^p, \mathbf{W}_{22}^p, \dots, \mathbf{W}_{LL}^p)$
and $\mathbf{W}^d = \operatorname{diag}(\mathbf{W}_{11}^d, \mathbf{W}_{22}^d, \dots, \mathbf{W}_{MM}^d)$.

Let $\mathbf{W} = \mathbf{P}\mathbf{W}^d\mathbf{P}^\top\mathbf{W}^p$
be the composite convolution kernel,
then we have
\begin{align}
\mathbf{x}' = \mathbf{W}\mathbf{x},
\end{align}
which implies that an IGC block is equivalent
to a regular convolution
with the convolution kernel
being the product of two sparse kernels.


\subsection{Analysis}
\noindent\textbf{Wider than regular convolutions.}
Recall that
the kernel size in the primary group convolution
is $S$
and
the kernel size in the secondary group convolution
is $1$ ($=1\times 1$).
Considering a single spatial position,
the number of the parameters
(equivalent to the computation complexity if the feature map size is fixed) in an IGC block
is
\begin{align}
T_{igc} &= (L \cdot M \cdot M \cdot S
+ M \cdot L \cdot L)  \nonumber\\
&=G^2\cdot (S/L+1/M),
\label{eqn:pdgccomplexity}
\end{align}
where $G=ML$ is the width (the number of channels) of an IGC block.

\begin{table*}
\setlength{\tabcolsep}{3pt}
\caption{The widths of our interleaved group convolution block
for various numbers of primary partitions $L$ and secondary partitions $M$
under the roughly-equal number of parameters: (i) $\approx~4672$
and (ii) $\approx~17536$.
The kernel size $S$ of primary group convolution is $9=3 \times 3$.
The width $LM$ is the greatest when $L\approx 9M$:
(i) $28 \approx 3 \times 9$
and (ii) $41 \approx 5 \times 9$. }
\label{tab:widthwithvariousLM}
\resizebox{1\linewidth}{!}{
\centering
\begin{tabular}{c|ccccccccc|ccccccccccc}
\hline
& \multicolumn{9}{c|}{(i): \#params $\approx~4672$} & \multicolumn{11}{c}{(ii): \#params $\approx~17536$}\\
\hline
$L$ &  $ 1 $ & $ 2 $  & $ 3 $  & $ 5 $ & $ 6 $ &  $ 12 $ &  $\bf{ 28 }$ & $ 40 $ & $ 64 $ &  $ 1 $ & $ 2 $  & $ 4 $  & $ 12 $ & $ 14 $ &  $ 23 $ &   $ 28 $ & {$ \bf{41 }$} &$  64 $ & $ 85 $ & $ 128 $ \\
\hline
$M$ & $ 23 $ & $ 16 $  & $ 13 $  & $ 10 $ & $ 9 $ &  $ 6 $ &  {$\bf{ 3} $} & $ 2 $ & $ 1 $ & $ 44 $ & $ 31 $  & $ 22 $  & $ 12 $ & $ 11 $ & $ 8 $ &  $ 7 $ & $ \bf{5} $ & $ 3 $ & $ 2 $ & $ 1 $\\
\hline
\#params & $ 4784 $ & $ 4672 $ & $ 4680 $  & $ 4750 $ & $ 4698 $ &  $ 4752 $ &   $ \bf{4620} $ & $ 4640 $ & $ 4672 $ & $ 17468 $ & $ 17422 $ & $ 17776 $  & $ 17280 $ &$  17402 $ &  $ 17480 $ &   $ 17836 $ & {$ \bf{17630} $} & $ 17472 $ & $ 17510 $ & $ 17536 $\\
\hline
Width & $ 23 $ & $ 32 $ & $ 39 $  & $ 50 $ & $ 54 $ &  $ 72 $ &  {$ \bf{84} $} &  $ 80 $ & $ 64 $ & $ 44 $ & $ 63 $ & $ 88 $  & $ 144 $ & $ 154 $ &  $ 184 $ &  $ 196 $ &  {$ \bf{205} $} & $ 192  $& $ 170 $ & $ 128 $ \\
\hline
\end{tabular}}
\vspace{-.3cm}
\end{table*}

For a regular convolution
with the same kernel size $S$
and the input and output width being $C$,
the number of parameters is
\begin{align}
T_{rc} = C \cdot C \cdot S .
\end{align}
Given the same number of parameters,
$T_{igc} = T_{rc} = T$,
we have $C^2 = \frac{1}{S}T$,
and $G^2 = \frac{1}{S/L + 1/M} T$.
It is easy to show that
\begin{align}
G > C,~\textrm{when~}\frac{L}{L-1} < MS.
\end{align}
Considering the typical case $S = 3 \times 3$,
we have $G > C$ when $L>1$.
In other words, an IGC block is wider than a regular convolution,
except the extreme case that there is only one partition in primary group convolution ($L=1$).


\vspace{0.1cm}
\noindent\textbf{When is the widest?}
\label{sec:effectofLMOnWidth}
We discuss how the primary and secondary partition numbers
$L$ and $M$ affect the width.
Considering Equation~\ref{eqn:pdgccomplexity},
we have,
\begin{align}
  T_{igc} =&~L \cdot M \cdot M \cdot S + M \cdot L \cdot L \\
  =&~LM(MS + L) \\
  \geqslant&~LM \cdot 2 \sqrt{LMS} \\
  =&~2\sqrt{S}(LM)^{\frac{3}{2}} \\
  =&~2\sqrt{S}G^{\frac{3}{2}},
\end{align}
where the equality in the third line holds when $L=MS$.
It implies that
(i) given the number of parameters,
the width $G$ is upper-bounded,
\begin{align}
G  \leqslant \left(\frac{T_{igc}}{2 \sqrt{S}}\right)^{\frac{2}{3}}.
\end{align}
and (ii) when $L=MS$, the width is the greatest.

Table~\ref{tab:widthwithvariousLM}
presents
two examples.
We can see that when $L \approx 9M $ ($S=9$),
the width is the greatest:
$3\times 9\approx 28$ for \#params $\approx 4672$
and $5\times 9 \approx 41$ for \#params $\approx 17536$.


\vspace{0.1cm}
\noindent\textbf{Wider leads to better performance?}
We have shown that an IGC block is equivalent to
a single regular convolution,
with the convolution kernel composed from two sparse kernels:
$\mathbf{W} = \mathbf{P}\mathbf{W}^d\mathbf{P}^\top\mathbf{W}^p$.
Fixing the parameter number means the following constraint,
\begin{align}
\|\mathbf{W}^p\|_0 +
\|\mathbf{W}^d\|_0 = T,
\label{eqn:sparsityconstraint}
\end{align}
where $\|\cdot \|_0$ is an entry-wise $\ell_0$ norm of a matrix.
This equation means that
when the IGC is wider
(or the dimension of the input $\mathbf{x}$ is higher),
$\mathbf{W}^p$ and $\mathbf{W}^d$ are larger
but more sparse.
In other words,
the composite convolution kernel $\mathbf{W}$
is more constrained
as it becomes larger.
Consequently,
the increased width is probably
not fully explored and the performance might not be improved,
because of the constraint in the composite convolution kernel $\mathbf{W}$.
Our empirical results shown in Figure~\ref{fig:varyML} verify this point
and suggest that an IGC block near the greatest width,
e.g., $M=2$ in the two example cases in Figure~\ref{fig:varyML},
achieves the best performance.

\section{Discussions and Connections}
\label{sec:discussions}

We show that regular convolutions, summation fusion preceded by group convolution as studied in ResNeXt~\cite{XieGDTH16}
and the Xception block~\cite{Chollet16a} are special IGC blocks,
and discuss several possible extensions.

\begin{figure}
\centering
\subfigure[]{\includegraphics[scale=0.7,clip]{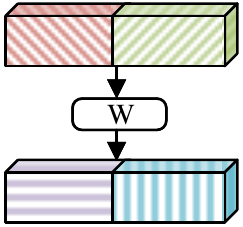}}~~~~~~~
\subfigure[]{\label{fig:regularconvPDGCview:PDGCView}\includegraphics[scale=0.7,clip]{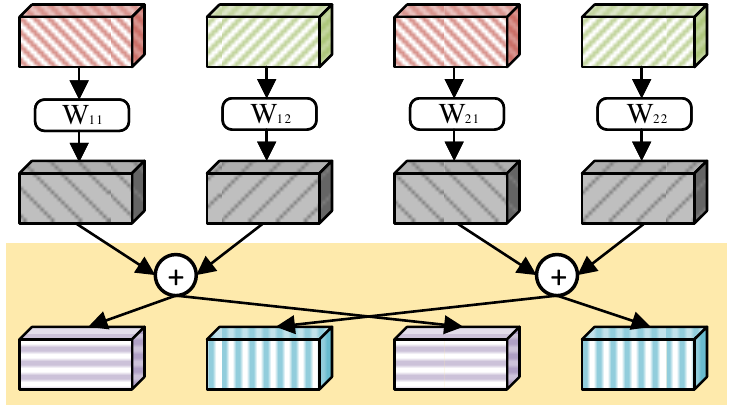}}
\caption{(a) Regular convolution. (b) Four-branch representation of the regular convolution.
The shaded part in (b), we call cross-summation,
is equivalent to a
three-step transformation: permutation, secondary group convolution,
and permutation back.}
\label{fig:regularconvPDGCview}
\vspace{-.3cm}
\end{figure}

\vspace{.1cm}
\noindent\textbf{Connection to regular convolutions.}
A regular convolution over a single spatial position
can be written as
$\mathbf{x}' = \mathbf{W}\mathbf{x}$,
where $\mathbf{x}$ is the input,
$\mathbf{W}$ is the weight matrix corresponding to the convolution kernel,
and $\mathbf{x}'$ is the output.
We show the equivalent IGC form
by taking $L=4$ as an example,
which is illustrated in Figure~\ref{fig:regularconvPDGCview}.
The general equivalence for other $L$ can be similarly derived.

Its IGC form is given as follows,
\begin{align}
\bar{\mathbf{x}}' =&
     \mathbf{P}\mathbf{W}^d
     \mathbf{P}^\top
     \mathbf{W}^p
     \bar{\mathbf{x}}.
     \label{eqn:regularconvolutionequivalence}
\end{align}
Here,
$\bar{\mathbf{x}} = [\mathbf{x}^\top~\mathbf{x}^\top]^\top$,
and $\bar{\mathbf{x}}' = [\mathbf{x}'^\top~\mathbf{x}'^\top]^\top$.
$\mathbf{W}^p$ is a block-diagonal matrix,
\begin{align}
\mathbf{W}^p = \operatorname{diag}(\mathbf{W}_{11},
\mathbf{W}_{12}, \mathbf{W}_{21}, \mathbf{W}_{22}).
\end{align}
$\mathbf{W}_{ij}$ is a block of $\mathbf{W}$
which is in the form of $2 \times 2$ blocks,
\begin{equation}
\mathbf{W} = \begin{bmatrix}
       \mathbf{W}_{11} & \mathbf{W}_{12} \\[0.3em]
       \mathbf{W}_{21} & \mathbf{W}_{22}
     \end{bmatrix}.
\end{equation}
$\mathbf{W}^d$ is a diagonal block matrix
with $M$ ($=$
half of the dimension of $\mathbf{x}$) blocks of size $L \times L$,
where $L=4$.
All block matrices in $\mathbf{W}^d$
are the same:
\begin{align}
\mathbf{W}_{11}^d = \mathbf{W}_{22}^d
=\dots = \mathbf{W}_{MM}^d =
\begin{bmatrix}
       1 & 1 & 0 & 0 \\[0.3em]
       0 & 0 & 1 & 1 \\[0.3em]
       1 & 1 & 0 & 0 \\[0.3em]
       0 & 0 & 1 & 1
\end{bmatrix}.
\end{align}

\vspace{.1cm}
\noindent\textbf{Connection to summation fusion.}
The summation fusion block~\cite{WangWZZ16} (like used in ResNeXt~\cite{XieGDTH16}),
is composed of a group of branches, e.g., $L$ convolutions\footnote{We discuss the case that each branch (partition) in summation fusion
includes only one convolutional layer. Our approach can also be extended
to more than one layer in each partition.}
(as defined in Equation~\ref{eqn:kgroupconvolution})
followed by a summation operation,
which is written as follows,
\begin{align}
\mathbf{x}' = \sum\nolimits_{i=1}^L \mathbf{y}_i,
\end{align}
where $\mathbf{x}'$
is the input of the next group convolution
in which the inputs of all the branches are the same.
Unlike the shaded part in Figure~\ref{fig:regularconvPDGCview:PDGCView} for regular convolution,
summation fusion receives all the four inputs
and sum them together
as the four outputs,
which are the same.

In the form of interleaved group convolutions,
the secondary group convolution is simple
and the kernel parameters in each convolution are all $1$,
i.e., the matrix $\mathbf{W}_{mm}^d$ in Equation~\ref{eqn:dualgroupconvolution}
is an all-one matrix.
For example,
in the case that there are $4$ primary partitions,
\begin{align}
\mathbf{W}_{11}^d = \mathbf{W}_{22}^d
=\dots = \mathbf{W}_{MM}^d =
\begin{bmatrix}
       1 & 1 & 1 & 1 \\[0.3em]
       1 & 1 & 1 & 1 \\[0.3em]
       1 & 1 & 1 & 1 \\[0.3em]
       1 & 1 & 1 & 1
\end{bmatrix}.
\end{align}

\vspace{.1cm}
\noindent\textbf{Xception is an extreme case.}
We discuss two extreme cases:
$L=1$ and $M=1$.
In the case where $L=1$,
the primary group convolution becomes a regular convolution,
and the secondary group convolution behaves like
assigning each channel with a different weight.

In the case where $M=1$,
the primary group convolution becomes an extreme group convolution:
a channel-wise group convolution,
and the secondary group convolution becomes a $1 \times 1$ convolution.
This extreme case is close to Xception
(standing for Extreme Inception)~\cite{Chollet16a}
that
consists of a channel-wise spatial convolution preceded
by a $1\times 1$ convolution\footnote{The similar idea is also studied in deep root~\cite{IoannouRCC16}.}.
It is pointed in~\cite{Chollet16a} that
performing the $1 \times 1$ convolution before or after the channel-wise spatial convolution
does not make difference.
Section~\ref{sec:effectofLMOnWidth}
shows that the two extreme cases do not lead to the greater width
except the trivial case that $L=9$ and $M=1$ ($L=9M$).
Our empirical results shown in Figure~\ref{fig:varyML} also indicate that
$L=1$ performs poorly
and $M=1$ performs well
but not the best.

\vspace{.1cm}
\noindent\textbf{Extensions and variants.}
First,
the convolution kernels in primary and secondary group convolutions
are changeable:
primary group convolution uses $1\times 1$ convolution kernels
and secondary group convolution uses spatial (e.g., $3\times 3$) convolution kernels.
Our empirical results show that
such a change does not make difference.
Second,
secondary group convolution can be replaced by a linear projection,
or a $1\times 1$ convolution,
which also blends the channels across partitions
outputted by primary group convolution.
This results in a network like discussed in~\cite{Chollet16a, IoannouRCC16}.
Secondary group convolutions can also adopt spatial convolutions.
Both are not our choice
because extra parameters and computation complexity are introduced.

Last,
our approach appears
to be complementary to existing methods.
Other spatial convolutional kernels, such as $3 \times 1$
and $1 \times 3$, can also be used in our primary group convolutions:
decompose a $3 \times 3$ kernel into two successive kernels,
$3 \times 1$ and $1 \times 3$.
The number of output channels of primary group convolution
can also be decreased, which is like a bottleneck design.
These potentially further improve the parameter efficiency.

\begin{table*}
\caption{The architectures
of networks with regular convolutions (RegConv-$Wc$ with $c$
being the channel number (width) at the first stage),
with summation fusions (SumFusion),
and with interleaved group convolutions (IGC-$L4M8$, IGC-$L24M2$, IGC-$L32M26$).
$B$ is the number of blocks at each stage.
$4 \times (3 \times 3, 8)$ means
a group convolution with $4$ partitions,
with the convolution kernel on each partition being
$(3\times 3, 8)$.}
\label{tab:architecture}
\centering

\renewcommand{\arraystretch}{1.2}
\footnotesize
  \begin{tabular}{c|c|c|c|c|c}
  \hline
      Output size  &  SumFusion & RegConv-$Wc$ &  IGC-$L4M8$  & IGC-$L24M2$ & IGC-$L32M26$\\
      \hline
      $32\times 32$  &  $(3\times 3,8)$  & $(3\times 3,c)$ & $(3\times 3,32)$ & $(3\times 3, 48)$   & $(3 \times 3, 26 \times 32)$ \\
    \hline
     $32\times 32$   & $ \begin{bmatrix}
     4 \times (3 \times 3 , 8) \\[0.3em]
     \operatorname{Summation}
     \end{bmatrix} \times B$
    & $(3\times 3, c)\times B$
    & $ \begin{bmatrix}
         4 \times (3 \times 3, 8) \\[0.3em]
         8 \times (1 \times 1, 4)
         \end{bmatrix} \times B$
    &
    $ \begin{bmatrix}
     24 \times (3 \times 3, 2) \\[0.3em]
     2 \times (1 \times 1, 24)
     \end{bmatrix} \times B$
     &
     $ \begin{bmatrix}
     32 \times (3 \times 3, 26) \\[0.3em]
     26 \times (1 \times 1, 32)
     \end{bmatrix} \times 6$
     \\
    \hline
         $16\times 16$   & $ \begin{bmatrix}
     4 \times (3 \times 3, 16) \\[0.3em]
     \operatorname{Summation}
     \end{bmatrix} \times B$ &
     $(3\times 3,2c ) \times B$
        & $ \begin{bmatrix}
             4 \times (3 \times 3, 16) \\[0.3em]
             16 \times (1 \times 1, 4)
             \end{bmatrix} \times B$
        & $ \begin{bmatrix}
     24 \times (3 \times 3, 4) \\[0.3em]
     4 \times (1 \times 1, 24)
     \end{bmatrix} \times B$
     &
     $ \begin{bmatrix}
     32 \times (3 \times 3, 52) \\[0.3em]
     52 \times (1 \times 1, 32)
     \end{bmatrix} \times 6$

                     \\
    \hline
        $8\times 8$   & $ \begin{bmatrix}
     4 \times (3 \times 3, 32) \\[0.3em]
     \operatorname{Summation}
     \end{bmatrix} \times B$ & $ (3\times 3, 4c) \times B$
       & $ \begin{bmatrix}
            4 \times (3 \times 3, 32) \\[0.3em]
            32 \times (1 \times 1, 4)
            \end{bmatrix} \times B$
                     & $ \begin{bmatrix}
     24 \times (3 \times 3, 8) \\[0.3em]
     8 \times (1 \times 1, 24)
     \end{bmatrix} \times B$
     &
     $ \begin{bmatrix}
     32 \times (3 \times 3, 104) \\[0.3em]
     104 \times (1 \times 1, 32)
     \end{bmatrix} \times 6$
     \\
      \hline
             $1\times 1$  & \multicolumn{5}{c}{average pool, fc, softmax}  \\
      \hline
      Depth & \multicolumn{4}{c|}{$3B +2$} & 20
     \\
      \hline
  \end{tabular}
\end{table*}

\begin{table*}
	\centering
	\caption{The number of parameters
		of networks used in our experiments
		and the computation complexity
		in terms of FLOPs (\# of multiply-adds).
		The statistics of the summation fusion networks are nearly the same with RegConv-$W16$
		and are not included.
		For IGC-$L24M2$, the numbers of
		parameters are the smallest,
		and the computation complexities are the lowest.}
	\label{tab:networkparameters}
	\scriptsize
	\begin{tabular}{c|c|c|c|c|c|c|c|c}
		\hline
		\multirow{2}{*}{D} &\multicolumn{4}{c|}{\#Params ($\times$M)} & \multicolumn{4}{c}{FLOPs ($\times 10^8$)}\\
		\cline{2-9}
		&   RegConv-$W16$ & RegConv-$W18$ & IGC-$L4M8$ & IGC-$L24M2$ &   RegConv-$W16$ & RegConv-$W18$ & IGC-$L4M8$ & IGC-$L24M2$\\
		\hline
		$8$&  $0.075$	& $0.095$	& $0.078$ &	$\mathbf{0.047}$ & $0.122$	& $0.154$	& $0.131$ &	$\mathbf{0.099}$ \\
		$20$ & $0.27$	& $0.34$ & 	$0.27$ &	$\mathbf{0.15}$ & $0.406$	& $0.513$ & 	$0.424$ &	$\mathbf{0.288}$\\
		$38$ &  $0.56$	& $0.71$	& $0.57$	& $\mathbf{0.31}$ & $0.830$	& $1.05$	& $0.862$	& $\mathbf{0.571}$\\
		$62$ &  $0.95$	& $1.20$	& $0.96$	& $\mathbf{0.52}$ & $1.40$	& $1.77$	& $1.45$	& $\mathbf{0.948}$\\
		$98$ &  $1.53$	& $1.93$	& $1.56$	& $\mathbf{0.83}$ & $2.25$	& $2.84$	& $2.32$	& $\mathbf{1.51}$\\
		\hline
	\end{tabular}
	\vspace{-.3cm}
\end{table*}

\begin{table}
	\caption{Classification accuracy
		comparison on CIFAR-$ 10 $ and CIFAR-$ 100 $
		of the convolutional networks with regular convolutions (RegConv-$W16$, RegConv-$W18$),
		with summation fusions (SumFusion),
		and with interleaved group convolutions (IGC-$L4M8$, IGC-$L24M2$).
		The architecture description and the parameter number statistics
		are given in Table~\ref{tab:architecture}
		and in Table~\ref{tab:networkparameters}.}
	\label{tab:resultsWithRegular}
	\centering
	\scriptsize
	\resizebox{1\linewidth}{!}{
		\begin{tabular}{l|ccccc}
			\hline
			& SumFusion &  RegConv-$W16$ & RegConv-$W18$ & IGC-$L4M8$ & IGC-$L24M2$  \\
			\hline
			D & \multicolumn{5}{c}{CIFAR-$ 10 $}  \\
			\hline
			$8$       &  $84.94\pm0.40  $ & $89.46\pm0.16  $ & $  90.30\pm0.25$ & $ 89.89\pm0.24 $ & $\mathbf{90.31\pm0.39}$\\
			$20$       & $ 88.71\pm0.46 $ & $  92.24\pm0.17 $ & $ 92.55\pm0.14$ & $92.54\pm0.37$ & $\mathbf{92.84\pm0.26}$ \\
			$38$         & $ 86.95\pm0.77 $ & $ 90.77\pm0.23$ & $91.57\pm0.09$ &$92.05\pm0.76$ & $\mathbf{92.24\pm0.62}$\\
			$62$  &  $82.66\pm0.75$ & $88.22\pm0.91 $ & $88.60\pm0.49$ &$89.23\pm0.89$ & $\mathbf{90.03\pm0.85}$ \\
			\hline
			D & \multicolumn{5}{c}{CIFAR-$ 100 $}  \\
			\hline
			$8$       &  $ 52.01\pm0.77 $ & $62.83\pm0.32  $ & $ 64.70\pm0.27$ & $64.18\pm0.70 $ & $\mathbf{65.60\pm0.59}$\\
			$20$       & $59.33\pm0.86 $ & $  67.90\pm0.14 $ & $ 68.71\pm0.32$ & $69.45\pm0.69$  & $\mathbf{70.54\pm0.61}$\\
			$38$         & $ 57.18\pm1.21 $ & $ 64.04\pm0.42$ & $65.00\pm0.57$ &$67.33\pm0.48$ & $\mathbf{69.56\pm0.76}$\\
			$62$  &  $48.68\pm3.84$ & $56.88\pm1.16 $ & $58.52\pm2.31$ &$63.06\pm1.42$ & $\mathbf{65.84\pm0.75}$\\
			\hline
		\end{tabular}
	}
	\vspace{-.3cm}
\end{table}

\section{Experiments}
\subsection{Datasets.}
\noindent\textbf{CIFAR.}
The CIFAR datasets~\cite{Alex2009}, CIFAR-$10$ and CIFAR-$100$,
are subsets of the $80$ million tiny images~\cite{TorralbaFF08}.
Both datasets contain $60000$ $32\times32$ color images with $50000$ images for training and $10000$ images for test.
The CIFAR-$10$ dataset has $10$ classes containing $6000$ images each.
There are $5000$ training images and $1000$ testing images per class.
The CIFAR-$100$ dataset has $ 100 $ classes containing $ 600 $ images each.
There are $500$ training images and $100$ testing images per class.
The standard data augmentation scheme we adopt is widely used for
this dataset~\cite{HeZRS16, HuangSLSW16, LeeXGZT15, HuangLW16a, LarssonMS16a, LinCY13, RomeroBKCGB14, SpringenbergDBR14, SrivastavaGS15}:
we first
zero-pad the images with $4$ pixels on each side, and then
randomly crop them to produce $32\times32$ images, followed by
horizontally mirroring half of the images.
We normalize the
images by using the channel means
and standard deviations.

\vspace{.1cm}
\noindent\textbf{SVHN.}
The Street View House Numbers (SVHN) dataset\footnote{\href{url}
{http://ufldl.stanford.edu/housenumbers/}}
is obtained from house numbers in Google Street View images.
SVHN contains $73,257$
training images, $26,032$ test images, and
$531,131$ images as additional training.
Following~\cite{HuangSLSW16, LeeXGZT15, LinCY13},
we select out $400$
samples per class from the training set and $200$ samples from the additional set,
and use the remaining
images as the training set without
any data augmentation.

\begin{figure*}[t]
\centering
\footnotesize
(a)~\includegraphics[width=.46\linewidth, clip]{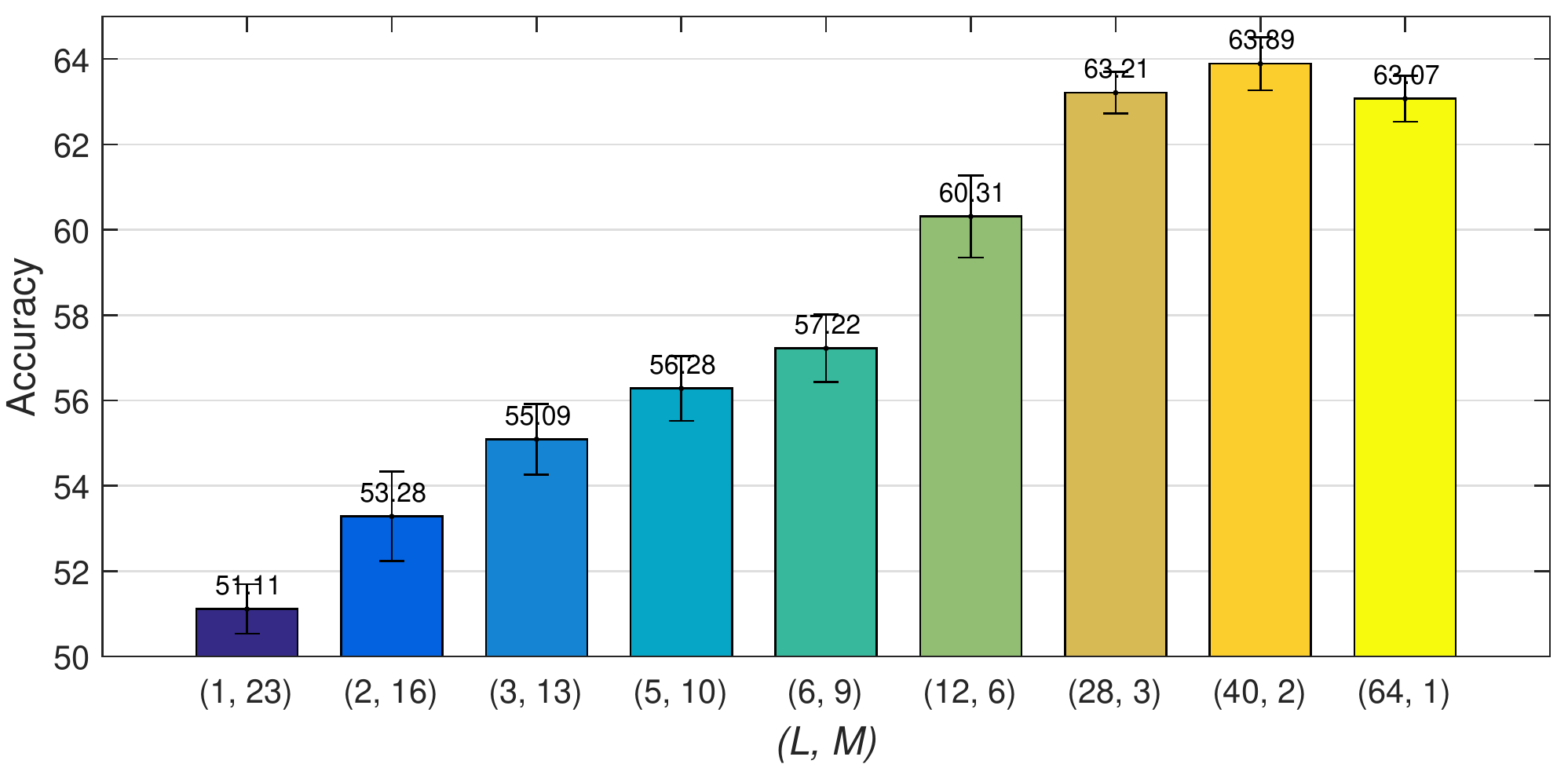}~~
(b)~\includegraphics[width=.46\linewidth, clip]{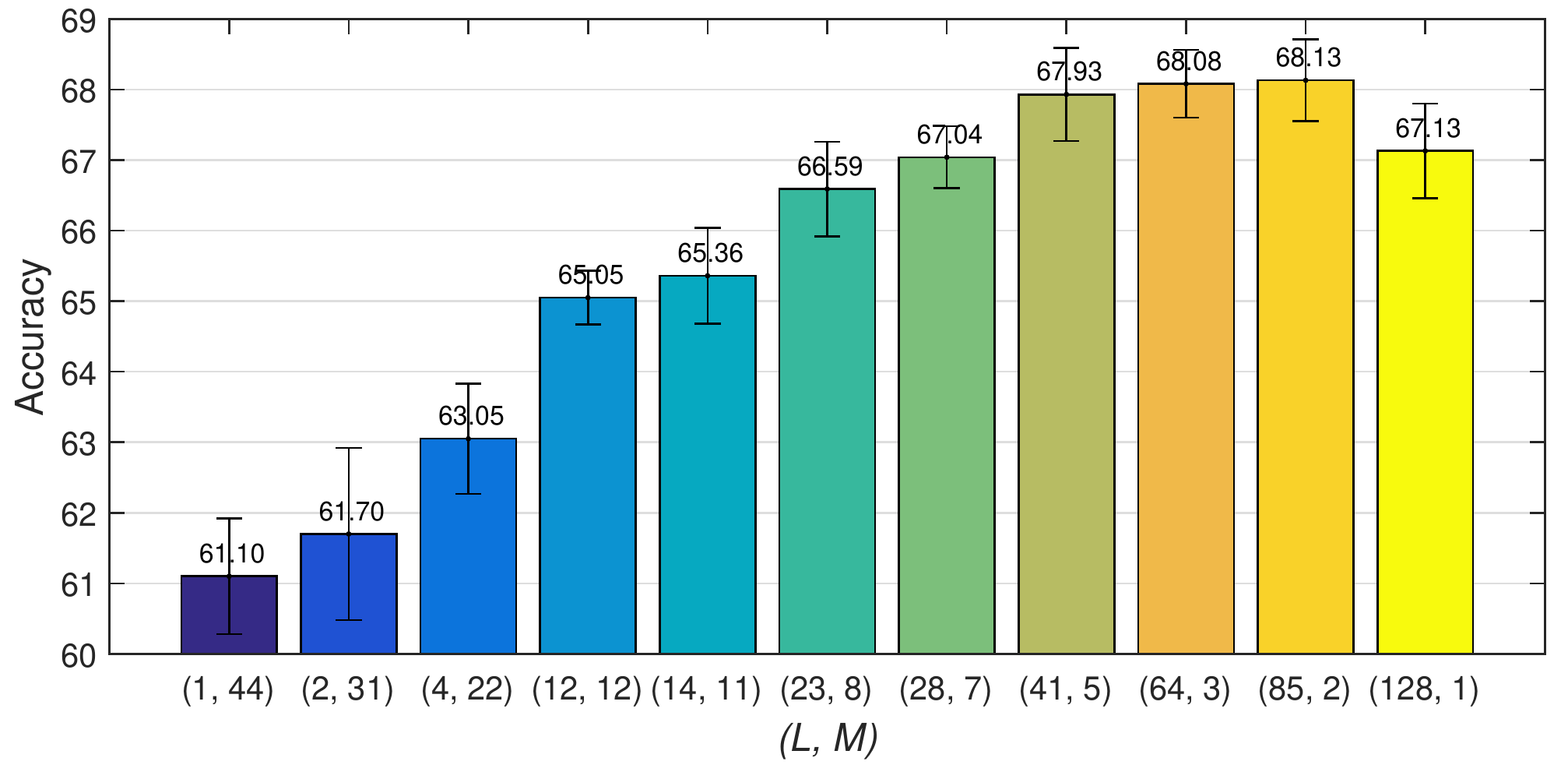}
   \caption{Illustrating the performances
   under different primary and secondary partition numbers $L$ and $M$
with same \#params
on CIFAR-$ 100 $.
We report the mean and the standard deviation over five runs.
(a) corresponds to (i) in Table~\ref{tab:widthwithvariousLM}
and (b) corresponds to (ii) with more parameters.}
\label{fig:varyML}
\end{figure*}

\subsection{Implementation Details}
We adopt batch normalization (BN)~\cite{IoffeS15} right
after each IGC block\footnote{There is no activation
between primary and secondary group convolutions.
Our experimental results
show that adding a nonlinear activation
between them deteriorates the classification performance.}
and
before nonlinear activation,
i.e., IGC + BN + ReLU.
We use the SGD algorithm with the Nesterov momentum,
and train all networks from scratch.
We initialize the weights
similar to~\cite{HeZRS15, HeZRS16,HuangLW16a},
and set the weight decay as $0.0001$ and the momentum as $0.9$\footnote{We did not
attempt to tune the hyper-parameters
for our networks,
and the chosen parameters may be suboptimal.}.

On CIFAR-$10$ and CIFAR-$100$,
we train all the models for $400$ epochs, with a total mini-batch
size $64$ on two GPUs.
The learning rate starts with $0.1$
and is reduced by a factor $10$ at the $200$, $300$, $350$ training epochs.
On SVHN, we train $40$ epochs for all the models,
with a total mini-batch size $64$ on two GPUs.
The learning rate starts with $0.1$
and is reduced by a factor $10$ at the $20$, $30$, $35$ training epochs.
Our implementation
is based on MXNet~\cite{ChenLLLWWXXZZ15}.


\begin{table*}
	\setlength{\tabcolsep}{8pt}
	\caption{Illustrating that our approach benefits from
		identity mappings.
		Classification accuracy
		comparison on CIFAR-$10$ and CIFAR-$100$ between
		ResNets
		and our approach with identity mappings.
Our network, IGC-$L24M2$+Ident. with fewer parameters and lower computation complexity (see Table~\ref{tab:networkparameters}),
performs the best.}
	\label{tab:resultsWithResidual}
	\centering
	\footnotesize
	{
		\begin{tabular}{l|cccc}
			\hline
			& RegConv-$W16$ + Ident. & RegConv-$W18$ + Ident. &  IGC-$L4M8$ +Ident.& IGC-$L24M2$ +Ident.
			\\
			\hline
			
			Depth&\multicolumn{4}{c}{CIFAR-$ 10 $}  \\
			\hline
			$ 50 $      &  $ 94.40\pm0.45  $ & $94.67\pm0.25$ & $94.74\pm0.54  $ & $\mathbf{94.88\pm0.32}$ \\
			$ 74 $       & $ 94.66\pm0.30 $ & $94.77\pm0.59$& $ 94.79\pm0.40 $ &  $\mathbf{94.95\pm0.23}$\\
			$ 98 $        & $ 94.71\pm0.44  $ & $94.95\pm0.39$& $ 94.81\pm0.30$ &  $\mathbf{95.15\pm0.48}$  \\
			\hline
			Depth &\multicolumn{4}{c}{CIFAR-$ 100 $}  \\
						\hline
		$ 50 $      &  $ 72.98\pm0.75 $ & $73.97\pm0.49$ & $74.00\pm0.69  $ & $\mathbf{74.89\pm0.67}$ \\
		$ 74 $       & $74.04\pm0.62 $ & $74.55\pm0.89$& $  75.15\pm0.49 $ &   $\mathbf{75.41\pm0.75}$  \\
		$ 98 $         & $ 74.49\pm0.66 $ & $75.30\pm0.88$ & $ 75.58\pm0.80$ & $\mathbf{76.15\pm0.50}$ \\
		\hline
		\end{tabular}
	}
\vspace{-.3cm}
\end{table*}

\begin{table}
	\caption{Imagenet classification results
		of a ResNet of depth $18$ and our approach.
		The network structure for ResNet can be found in~\cite{HeZRS16}.
Both ResNets and our networks contain four stages,
and when down-sampling is performed,
the channel number is doubled.
For ResNets, $C$ is the channel number at the first stage.
For our networks except IGC-$L100M2$+Ident., we double the channel number
by doubling $M$ and keeping $L$ unchanged.
For IGC-$L100M2$+Ident., we double the channel number
by doubling $L$ and keeping $M$ unchanged.
}
	\label{tab:imagenetresults}
	\centering
	\scriptsize
\resizebox{1\linewidth}{!}
{
	\begin{tabular}{l|c|c|cc|cc}
		\hline
		& {\#Params}&
		{FLOPs} & \multicolumn{2}{c|}{training error} & \multicolumn{2}{c}{testing error} \\
		\cline{4-7}
		& ($\times$M)& ($\times 10^9$) &top-1 & top-5 &  top-1 & top-5\\
		\hline
		ResNet ($C=64$) & $11.151$ & $1.8$ &$22.41$& $6.53$ & $31.06$& $11.38$  \\
		ResNet ($C=69$) & $11.333$& $2.1$& $21.43$& $5.96$ & $30.58$& $10.77$  \\
		IGC-$L4M32$+Ident. &$11.205$& $1.9$& $21.71$& $6.21$& $30.77$ & $10.99$ \\
		IGC-$L16M16$+Ident. &$11.329$&$2.2$& ${19.97}$& ${5.44}$& ${29.40}$ & ${10.32}$ \\
        IGC-$L100M2$+Ident. & $8.61$ & $1.3$ & $\mathbf{13.93}$ & $\mathbf{2.75}$ & $\mathbf{26.95}$ & $\mathbf{8.92}$ \\
		\hline
	\end{tabular}
}
\vspace{-.5cm}
\end{table}

\begin{table*}[t]
	\setlength{\tabcolsep}{12pt}
	\centering
	\caption{
		Classification error comparison
		with the state-of-the-arts. The best, second-best, and third-best accuracies are highlighted in red, green, and blue.
	}
	\label{tab:comparisonwithstateofthearts}%
	\footnotesize
	\begin{tabular}{@{\extracolsep{\fill}}l|c | c | c | c | c}
		\hline
		& Depth & \#Params & CIFAR-$10$ & CIFAR-$100$ & SVHN\\
		\hline
		Network in Network~\cite{LinCY13} &  -    &  -    & $8.81$  & $35.68$     & $2.35$\\
		All-CNN~\cite{SpringenbergDBR14} &  -    &  -    & $7.25$  & $33.71$ & - \\
		FitNet~\cite{RomeroBKCGB14} & -     & -     & $8.39$  & $35.04$ & $2.42$ \\
		Deeply-Supervised Nets~\cite{LeeXGZT15} &  -    &  -    & $8.22$  & $34.57$ & $1.92$ \\
		\hline
		Swapout~\cite{SinghHF16} &$ 20$    & $1.1$M & $6.58$  & $25.86$ & -\\
		&$ 32$    & $7.4$M & $4.76$  & $22.72$ & -\\
		\hline
		Highway~\cite{SrivastavaGS15} &  -    &  -    & $7.72$  & $32.39$ & -\\
		\hline
		DFN~\cite{WangWZZ16} &$ 50$    & $3.7$M & $6.40$  & $27.61$ & -\\
		&$ 50$    & $3.9$M & $6.24$  & $27.52$ & -\\
		\hline
		FractalNet~\cite{LarssonMS16a} &$ 21$    &  $38.6$M & $5.22$  & $23.30$ & $2.01$\\
		With dropout \& droppath &$ 21$    &  $38.6$M & $4.60$  & $23.73$ & $1.87$\\
		\hline
		ResNet~\cite{HeZRS16} & $110$   &  $1.7$M & $6.61$  & -     & -\\
		ResNet~\cite{HuangSLSW16}  & $110$   &  $1.7$M & $6.41$  & $27.76$ & $1.80$\\
		ResNet (pre-activation)~\cite{HeZRS16ECCV} & $164$   &  $1.7$M & $5.46$  & $24.33$ & -\\
		& $1001$  &  $10.2$M & $4.92$  & $22.71$ & -\\
		\hline
		ResNet with stochastic depth~\cite{HuangSLSW16} & $110$   &  $1.7$M & $5.25$  & $24.98$ & $1.75$\\
		& $1202$  &  $10.2$M & $4.91$  & -     & -\\
		\hline
		Wide ResNet~\cite{ZagoruykoK16} &$ 16$    &  $11.0$M & $4.27$  & $20.43$ & -\\
		&$ 28$    &  $36.5$M & $4.00$  & $19.25$ & - \\
		With dropout  &$ 16$    &  $2.7$M & -     & -     & $1.64$\\
		\hline
		RiR~\cite{TargAL16} & $18$   & $10.3$M& $5.01$  & $22.90$ & -\\
		\hline
		Multi-ResNet~\cite{AbdiN16} & $200$   & $10.2$M& $4.35$  & $20.42$ & -\\
		& $26$   & $72$M& $3.96$  & $19.45$     & -\\
		\hline
		DenseNet ($k=24$)~\cite{HuangLW16a}  & $100$   &  $27.2$M & $3.74$  & $19.25$ & {$1.59$}\\
		DenseNet-BC ($k=24$)~\cite{HuangLW16a}  & $250$   &  $15.3$M & $3.62$  & {\color{blue}$\mathbf{17.60}$} & $1.74$\\
		DenseNet-BC ($k=40$)~\cite{HuangLW16a}  & $190$   &  $25.6$M & $3.46$  & {\color{red}$\mathbf{17.18}$} & $-$\\
		\hline
		ResNeXt-29, $8\times 64$d~\cite{XieGDTH16} &$ 29$    & $34.4$M & ${3.65}$ & ${17.77}$  & $-$ \\
		ResNeXt-29, $16\times 64$d~\cite{XieGDTH16} &$ 29$    & $68.1$M & ${3.58}$ & {\color{green}$\mathbf{17.31}$}  & $-$ \\
		\hline
        DFN-MR1~\cite{ZhaoWLTZ16} &$ 56$    & $1.7$M & ${4.94}$ & ${24.46}$  & $1.66$ \\
		DFN-MR2~\cite{ZhaoWLTZ16} &$ 32$    & $14.9$M& $3.94$  & $19.25$ & {\color{red}$\mathbf{1.51}$} \\
		DFN-MR3~\cite{ZhaoWLTZ16} &$ 50$    & $24.8$M& $3.57$ & {$19.00$} & {\color{green}$\mathbf{1.55}$}\\
		\hline
		IGC-$L16M32$ & $20$ & $17.7$M& {\color{blue}$\mathbf{3.37}$} & $19.31$ & {$1.63$} \\
		IGC-$L450M2$ & $20$ & $19.3$M& {\color{red}$\mathbf{3.25}$} & ${19.25}$ & $-$ \\
		IGC-$L32M26$ & $20$ & $24.1$M& {\color{green}$\mathbf{3.31}$} & ${18.75}$ & {\color{blue}$\mathbf{1.56}$} \\
		\hline
	\end{tabular}%
\vspace{-.4cm}
\end{table*}%

\subsection{Empirical Study}

\noindent\textbf{Comparison with regular convolution and summation
	fusion.}
We compare five networks:
convolutional networks with regular convolutions
(RegConv-$W16$, RegConv-$W18$),
with summation fusions (SumFusion),
and with interleaved group convolution blocks (IGC-$L4M8$, IGC-$L24M2$).
Network architectures, parameter numbers and computation complexities
are given in Table~\ref{tab:architecture}
and in Table~\ref{tab:networkparameters}.

The comparisons on CIFAR-$10$ and CIFAR-$100$
are given in Table~\ref{tab:resultsWithRegular}.
It can be seen that
the overall performance of
our networks, IGC-$L4M8$,
are better than both RegConv-$W16$
containing slightly fewer parameters and RegConv-$W18$
containing more parameters,
demonstrating that our IGC block is more powerful
than regular convolutions.
Another model, IGC-$L24M2$,
containing much fewer parameters,
performs better than both RegConv-$W16$ and RegConv-$W18$.
The main reason lies in the advantage
that \emph{our IGC blocks increase the width
	and the parameters are exploited more efficiently}.
For example, on CIFAR-$100$, 
when the depth is $38$, IGC-$L4M8$
and IGC-$L24M2$ achieve $67.33\%$, $69.56\%$ accuracy,
about $2.3\%$, $4.5\%$ better than RegConv-$W18$.
The summation fusion (SumFusion) performs worse
because the summation fusion reduces the width
and the parameters are not very efficiently used.

\vspace{0.1cm}
\noindent\textbf{The effect of partition numbers.}
We have shown that
how the numbers of primary and secondary partitions
affect the width
and one extreme case of our approach
is Xception~\cite{Chollet16a}.
Now we empirically study how the performances
are affected by the partition numbers
and show that a typical setup, $M=2$,
performs better than
Xception~\cite{Chollet16a}.

To clearly show the effect,
we use networks
with $8$ layers:
$6$ IGC blocks, the first convolution layer,
and the last FC layer.
There is no down-sampling stage:
the map is always of size $32\times 32$.
We change the partition numbers,
$L$ and $M$,
to guarantee the model size (the computation complexity) almost the same.
We consider two cases for an IGC block:
(i) the parameter number ($9LM^2+ML^2$) is approximately $4672$
and (ii) the parameter number is approximately
$17536$ (see Table~\ref{tab:widthwithvariousLM}).

The results are presented in Figure~\ref{fig:varyML}.
It can be observed that
the accuracy increases when the number of primary partitions becomes larger
(the number of secondary partitions becomes smaller)
till it reaches some number
and then decreases.
In the two cases,
the performance with $M=2$ secondary partitions
is better than Xception.
For example, in case (i),
IGC with $L=40$ and $M=2$ gets
$63.89\%$ accuracy, about $0.8\%$ better than IGC with $L=64$ and $M=1$, which gets $63.07\%$ accuracy.
We believe that the performance in general is a concave function
with respect to $M$ (or $L$) under roughly the same number of parameters,
and the performance is not the best when $M=1$ (i.e., Xception~\cite{Chollet16a}) or $L=1$.

\vspace{0.1cm}
\noindent\textbf{Combination with identity mappings.}
We show that our IGCNet also benefits from identity mappings
and achieves superior performance over ResNets~\cite{HeZRS16}.
We compare two networks with regular convolutions,
RegConv-$W16$ and RegConv-$W18$,
with IGC-$L4M8$ and IGC-$L24M2$.
The residual branch consists of
two regular convolution layers for ResNets~\cite{HeZRS16}
and two IGC blocks for our networks.

The results are shown in Table~\ref{tab:resultsWithResidual}.
One can see that our approaches, IGC-$L4M8$ + Ident. and IGC-$L24M2$+Ident.,
do not suffer from training difficulty because of identity mappings.
IGC-$L4M8$+Ident.
performs better (e.g., about $1\%$ accuracy improvement on CIFAR-$100$
with slightly more parameters, see Table~\ref{tab:networkparameters}) than RegConv-$W16$ + Ident.,
and performs similar
(with smaller \#parameters
and computation complexity, see Table~\ref{tab:networkparameters}) to RegConv-$W18$+Ident..
In addition, IGC-$L24M2$+Ident.,
with fewer parameters and lower computation complexity (see Table~\ref{tab:networkparameters}),
performs better than both RegConv-$W16$+Ident. and RegConv-$W18$+Ident.,
which again demonstrates that our IGC block can exploit the parameters efficiently.

%

\vspace{0.1cm}
\noindent\textbf{ImageNet classification.}
We present the comparison to ResNets~\cite{HeZRS16} for ImageNet classification.
The ILSVRC $ 2012 $ classification dataset~\cite{DengDSLL009}
contains over $1.2$ million training images and $50,000$
validation images, and
each image is labeled from $1000$ categories.
We adopt the same data augmentation
scheme for the training images as in~\cite{HeZRS16,HeZRS16ECCV}.
The models are trained for $95$ epochs
with a total mini-batch size $256$ on $8$ GPUs.
The learning rate starts with $0.1$ and is reduced by a factor $10$
at the $30$, $60$, $90$ epochs.
A single $224\times 224$ center crop from an image is used to evaluate at test time.
Our purpose is not to push the state-of-the-art results,
but to demonstrate the powerfulness of our approach.
So we use the comparison to ResNet-$18$
as an example.

The result is depicted in Table~\ref{tab:imagenetresults}.
(i) Our approach,
IGC-$L4M32$+Ident., performs better
than ResNet ($C=64$) that contains slightly fewer parameters.
(ii) Our approach IGC-$L16M16$+Ident.
performs better than ResNet ($C=69$)
that has approximately the same number of parameters and computation complexity:
our model gets about $1.5\%$ reduction for top-$1$ error
and $1\%$ reduction for top-$5$ error.
(iii) Our approach IGC-$L100M2$+Ident.
gets the best result
with a much smaller number of parameters
and smaller computation complexity.
We also notice that the training error of
our approach
is smaller than ResNets,
suggesting that the gains are not from
regularization but from richer representation.

%
%
%
%

\subsection{Comparison with the State-of-the-Arts}
We compare our approach with the state-of-the-art algorithms.
The comparisons are reported in Table~\ref{tab:comparisonwithstateofthearts}.
We do not optimally tune the partition numbers
for our network
since
the NVIDIA CuDNN library does not
support group convolutions yet,
making the group convolution operation slow
in practical implementation.

Our networks
contain $20$ layers:
$18$ interleaved group convolution blocks, the first convolution layer and the last FC layer (see IGC-$L32M26$ in Table~\ref{tab:networkparameters} as an example.
We double the width by doubling $M$ when down-sampling the feature map at each stage).
The best, second-best, and third-best accuracies are highlighted in
red, green, and blue.
It can be seen that our networks achieve competitive performance:
the best accuracy on CIFAR-$10$,
and the third-best accuracy on SVHN (close to the second-best accuracy).
Our performance would be better
if our network also adopts the bottleneck design as in DenseNet-BC~\cite{HuangLW16a}
and ResNeXt~\cite{XieGDTH16}
or adopts more primary partitions.

\section{Conclusion}
In this paper,
we present a novel convolutional neural network architecture,
which addresses the redundancy problem
of convolutional filters
in the channel domain.
The main novelty lies in
an interleaved group convolution block:
channels in the same partition
in the secondary group convolution
come from different partitions
used in the primary group convolution.
Experimental results demonstrate
that our network
is efficient in parameter and computation.
{\small
\bibliographystyle{ieee}
\bibliography{PDGC}
}

%
\section*{Appendices}

\begin{table*}[t]
	\setlength{\tabcolsep}{3pt}
	\caption{Example configurations of GPCs
		for various numbers ($L$) of partitions and
        various numbers ($M$) of channels in each partition,
		under the roughly-equal number of parameters: (i) $\approx~4672$
		and (ii) $\approx~17536$.
		The kernel size $S$ in group convolutions is $9=3 \times 3$. }
	\label{tab:GPCwidthwithvariousLM}
	\resizebox{1\linewidth}{!}{
		\centering
		\begin{tabular}{c|cccccccc|c|cccccccccc|c}
			\hline
			& \multicolumn{9}{c|}{(i) \#params $\approx~4672$} & \multicolumn{11}{c}{(ii) \#params $\approx~17536$}\\
			\hline
			& \multicolumn{8}{c|}{GPC} & IGC & \multicolumn{10}{c|}{GPC} & IGC\\
			\hline
			$L$ &  $ 1 $ & $ 2 $  & $ 3 $  & $ 5 $ & $ 10 $ &  $ 19 $ &  $30$ & $ 64 $  & $40$ &  $ 1 $ & $ 2 $  & $ 3 $  & $ 6 $ & $ 11 $ &  $ 15 $ &   $ 18 $ & $29$ & $ 62 $ & $ 128 $ & $85$ \\
			\hline
			$M$ & $ 22 $ & $ 15 $  & $ 12 $  & $ 8 $ & $ 5 $ &  $ 3 $ &  $2$& $ 1 $ & $2$&  $ 42 $ & $ 28 $  & $ 22 $  & $ 14 $ & $ 9 $ & $ 7 $ &  $ 6 $ & $ 4 $  & $ 2 $ & $ 1 $ & $2$\\
			\hline
			\#params & $ 4840 $ & $ 4950 $ & $ 5184 $  & $ 4480 $ & $ 4750 $ &  $ 4788 $ &   $ 4680 $ & $ 4672 $ & $4640$ & $ 17640 $ & $ 17248 $ & $ 17424 $   &$  17820 $ &  $ 17640 $ &   $ 17496 $ & $ 17632 $ & $ 17640 $ & $ 17608 $ & $ 17532 $ & $17510$\\
			\hline
			Width & $ 22 $ & $ 30 $ & $ 36 $  & $ 40 $ & $ 50 $ &  $ 54 $ &  $ 60 $ &  $ 64 $ & $80$ & $ 42 $ & $ 56 $ & $ 66 $  & $ 84 $ & $ 99 $ &  $ 105 $ &  $ 108 $ &  $ 116 $ & $ 124 $ & $ 128 $ & $170$\\
			\hline
	\end{tabular}}
\end{table*}

\begin{figure*}[t]
	\centering
	\footnotesize
	(a)~\includegraphics[width=.46\linewidth, clip]{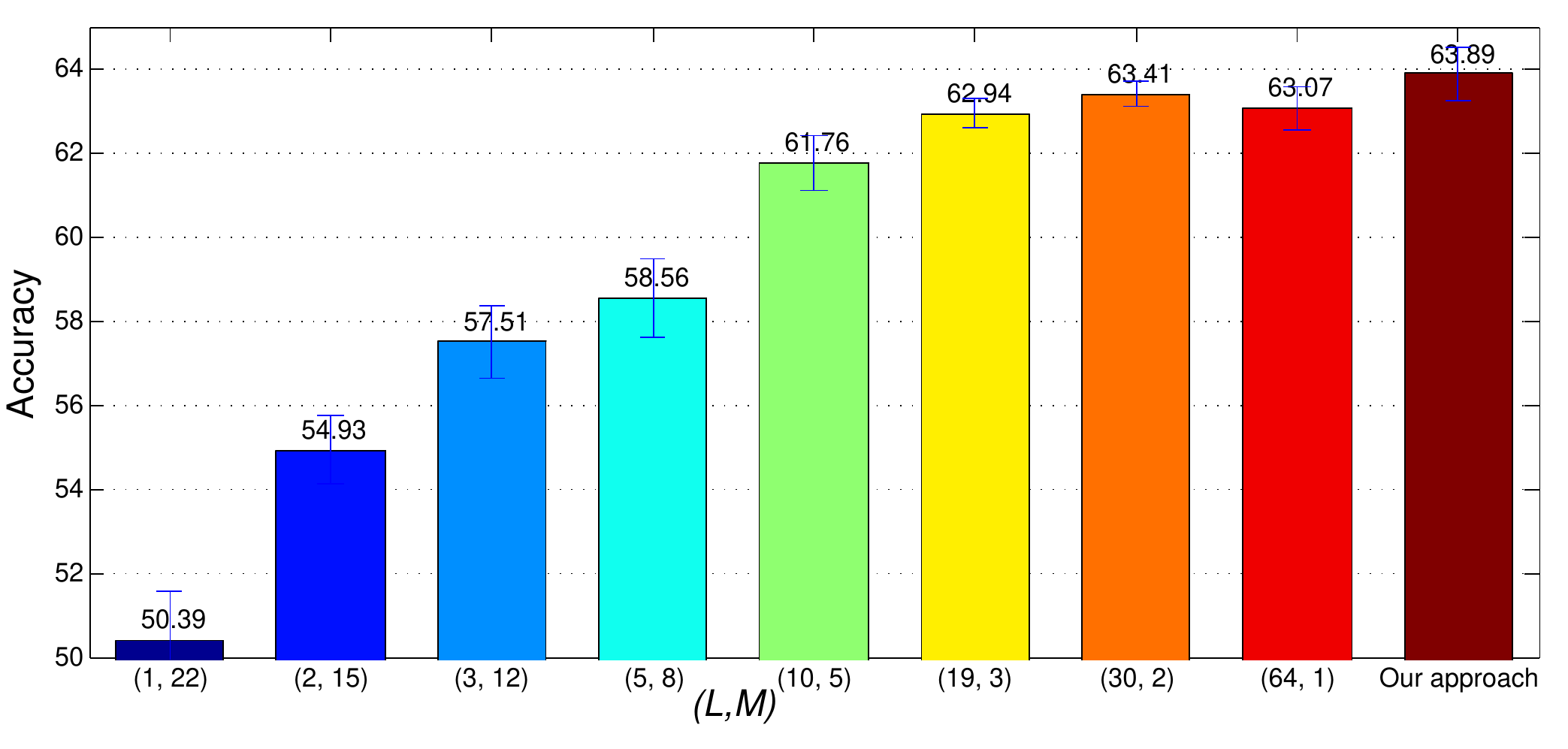}~~
	(b)~\includegraphics[width=.46\linewidth, clip]{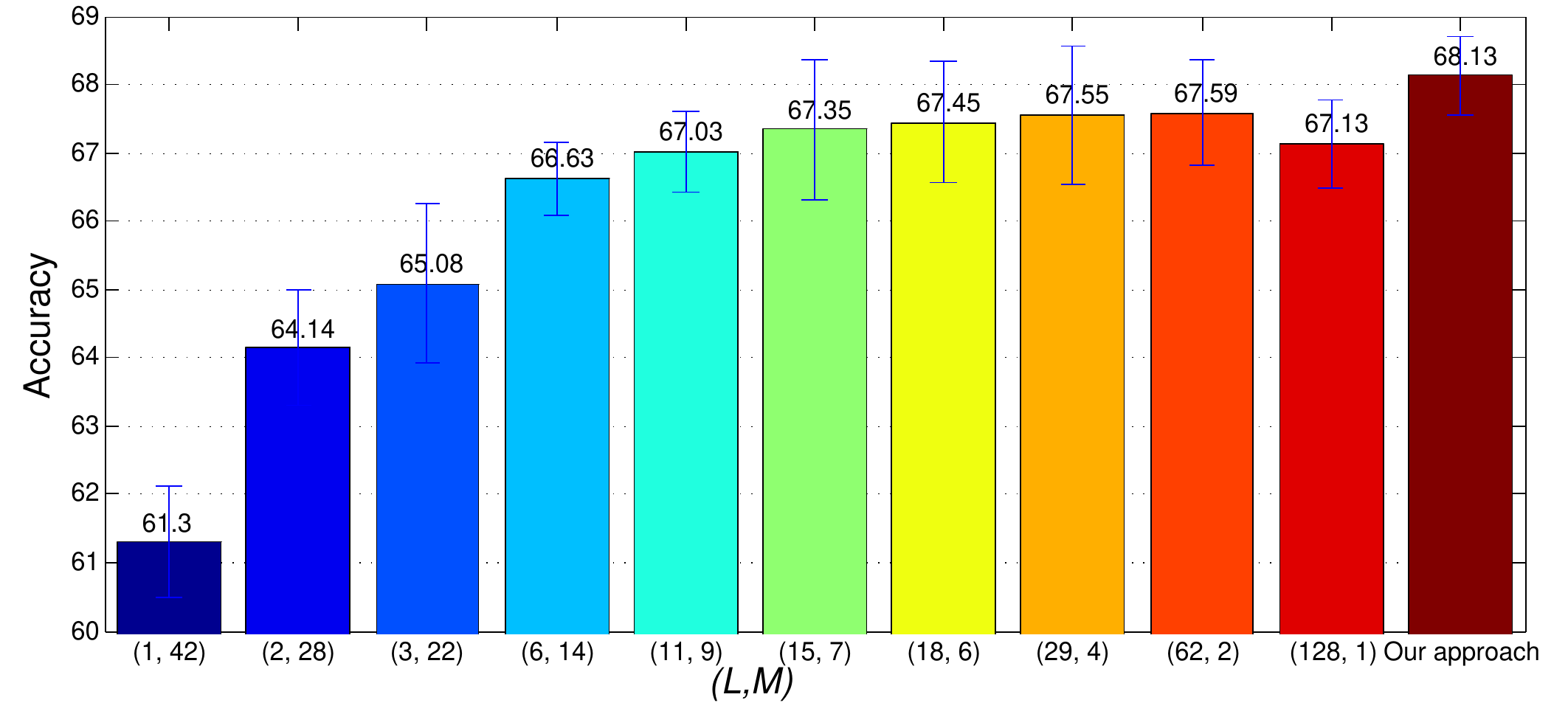}
	\caption{Illustrating the performances
		between our approach and the networks
stacking GPC with various numbers $L$ and $M$
		with same \#params
		on CIFAR-$ 100 $.
		We report the mean and the standard deviation over five runs.
		(a) corresponds to (i) in Table~\ref{tab:GPCwidthwithvariousLM}
		and (b) corresponds to (ii) with more parameters.}
	\label{fig:GPCvaryML}
\end{figure*}

\noindent
\textbf{Comparison with alternative structures.}
The proposed network is
a stack of interleaved group convolution (IGC) blocks,
where secondary group convolutions to blend the channels
across partitions outputted by primary group convolutions.

In the main paper (Extensions and variants, Section 4),
we discuss that a point-wise convolution,
i.e., a $1\times 1$ convolution,
as an alternative of
secondary group convolutions,
introduces extra parameters and computation complexity.
Such an alternative is also mentioned or discussed in
Xception~\cite{Chollet16a} and deep roots~\cite{IoannouRCC16}.
We denote this alternative block
as Group-and-Point-wise Convolution (GPC).
The number of parameters for one GPC block is
$L\cdot M \cdot M\cdot S + L\cdot M \cdot L\cdot M$,
where $L$ is the number of partitions in group convolution,
$M$ is the number of channels in each partition,
and thus
$LM$ is the number of total channels.

We replace IGC blocks in our networks
using GPC blocks,
with almost the same number of parameters.
Table~\ref{tab:GPCwidthwithvariousLM}
presents
the configurations:
(i) \#params $\approx 4672$
and (ii) \#params $\approx 17536$
(See Table 1 in the main paper
for the configurations of our IGC blocks).
The results,
together with our approach (with two secondary partitions) are presented in Figure~\ref{fig:GPCvaryML}
(More results about our networks are
shown in Figure 3 in the main paper).
We can see that our networks perform better
and achieve around $0.5\%$ improvement
in both cases.

\noindent
\textbf{More on the connection to regular convolutions.}
We rewrite Equation~\ref{eqn:regularconvolutionequivalence} as
the following,
\begin{align}
\bar{\mathbf{x}}' =&
     \mathbf{P}\mathbf{W}^d
     \mathbf{P}^\top
     \mathbf{W}^p
     \bar{\mathbf{x}}.
\end{align}
We discuss an extreme case:
primary group convolution is a channel-wise convolution.
Let $$\bar{\mathbf{x}} = [\mathbf{x}^\top \mathbf{x}^\top~\dots~\mathbf{x}^\top]^\top.$$
be formed by concatenating $\mathbf{x}$ $C$ times.
The primary group convolution is given as follows,
\begin{align}
\mathbf{W}^p = \operatorname{diag}(&\mathbf{w}_{11}^\top, \mathbf{w}_{12}^\top, \dots, \mathbf{w}_{1C}^\top, \\
&\mathbf{w}_{21}^\top, \mathbf{w}_{22}^\top, \dots, \mathbf{w}_{2C}^\top, \\
& ~~~~~~~~~~~~~\dots, \\
& \mathbf{w}_{C1}^\top, \mathbf{w}_{C2}^\top, \dots, \mathbf{w}_{CC}^\top),
\end{align}
where $\mathbf{w}_{ij}$
is a vector of $3 \times 3$.
The secondary group convolution is given as follows,
\begin{align}
\mathbf{W}^d =
\begin{bmatrix}
      \mathbf{1}^\top & \mathbf{0}^\top & \mathbf{0}^\top & \mathbf{0}^\top \\[0.3em]
       \mathbf{0}^\top & \mathbf{1}^\top & \mathbf{0}^\top & \mathbf{0}^\top \\[0.3em]
       \dots & \dots & \dots & \dots \\[0.3em]
       \mathbf{0}^\top & \mathbf{0}^\top & \dots & \mathbf{1}^\top \\[0.3em]
       \vdots & \vdots & \vdots & \vdots \\[0.3em]
       \mathbf{1}^\top & \mathbf{0}^\top & \mathbf{0}^\top & \mathbf{0}^\top \\[0.3em]
       \mathbf{0}^\top & \mathbf{1}^\top & \mathbf{0}^\top & \mathbf{0}^\top \\[0.3em]
       \dots & \dots & \dots & \dots \\[0.3em]
       \mathbf{0}^\top & \mathbf{0}^\top & \dots & \mathbf{1}^\top \\[0.3em]
\end{bmatrix},
\end{align}
where $\mathbf{1}$ is a vector of $C$ ones,
$\mathbf{0}$ is a vector of $C$ zeros.
$\mathbf{W}^d$ consists of $C \times 1$ blocks,
with each block being a matrix of $C \times C^2$.
Thus, we have
$$\bar{\mathbf{x}} = [\mathbf{x}'^\top \mathbf{x}'^\top~\dots~\mathbf{x}'^\top]^\top.$$

\end{document}